\documentclass{bmvc2k}

\title{Orientation-boosted Voxel Nets for 3D Object Recognition}

\addauthor{Nima Sedaghat}{nima@cs.uni-freiburg.de}{1}
\addauthor{Mohammadreza Zolfaghari}{zolfagha@cs.uni-freiburg.de}{1}
\addauthor{Ehsan Amiri}{amirie@cs.uni-freiburg.de}{1}
\addauthor{Thomas Brox}{brox@cs.uni-freiburg.de}{1}

\addinstitution{
Computer Vision Group\\ 
University of Freiburg\\
Germany
}

\runninghead{\scriptsize Sedaghat \MakeLowercase{\etal}}{\scriptsize Orientation-boosted
Voxel Nets for 3D Object Recognition -- BMVC 2017}

\def\etal{\emph{et al}\bmvaOneDot}

\usepackage{graphicx}
\usepackage{amsmath,amssymb} 
\usepackage{color}
\usepackage{colortbl}
\usepackage{caption}
\usepackage{subcaption}
\usepackage{multirow}
\usepackage{bigstrut}
\usepackage{hhline}
\usepackage{gensymb}
\usepackage{booktabs}
\usepackage{mdframed}
\usepackage{substr}
\usepackage{breqn}
\usepackage{xspace}
\usepackage{xcolor}
\usepackage{gensymb}
\usepackage{siunitx}
\usepackage[export]{adjustbox}
\usepackage{float}

\definecolor{ncolor}{RGB}{0,180,0}

\DeclareMathOperator*{\argmax}{arg\,max}
\newcommand{\TR}[1]{{\color{ncolor}{#1}}}
\newcommand{\FA}[1]{{\color{red}{#1}}}

\begin{document}
\maketitle

\begin{abstract}
Recent work has shown good recognition results in 3D object recognition using 3D convolutional networks.
In this paper, we show that the object orientation plays an important role in 3D recognition.
More specifically, we argue that objects induce different features in the network under rotation.
Thus, we approach the category-level classification task as a multi-task problem, in which the network is trained to predict the pose of the object in addition to the class label as a parallel task. We show that this yields significant improvements in the classification results.
We test our suggested architecture on several datasets representing various 3D data sources: LiDAR data, CAD models, and RGB-D images. We report state-of-the-art results on classification as well as significant improvements in precision and speed over the baseline on 3D detection.
\end{abstract}

\section{Introduction}

Various devices producing 3D point clouds have become widely applicable in recent years, e.g., range sensors in cars and robots or depth cameras like the Kinect. Structure from motion and SLAM approaches have become quite mature and generate reasonable point clouds, too.
With the rising popularity of deep learning, features for recognition are no longer designed manually but learned by the network. Thus, moving from 2D to 3D recognition requires only small conceptual changes in the network architecture~\cite{wu_3D_2015,maturana_voxnet_2015}.

In this work, we elaborate on 3D recognition using 3D convolutional networks, where we focus on the aspect of auxiliary task learning. Usually, a deep network is directly trained on the task of interest, i.e., if we care about the class labels, the network is trained to produce correct class labels.
There is nothing wrong with this approach. However, it requires the network to learn the underlying concepts, such as object pose, that generalize to variations in the data. Oftentimes, the network does not learn the full underlying concept but some representation that only partially generalizes to new data.

\begin{figure}[]
  \begin{center}
    \includegraphics[trim={0 1cm 0 0},clip,width=.8\linewidth]{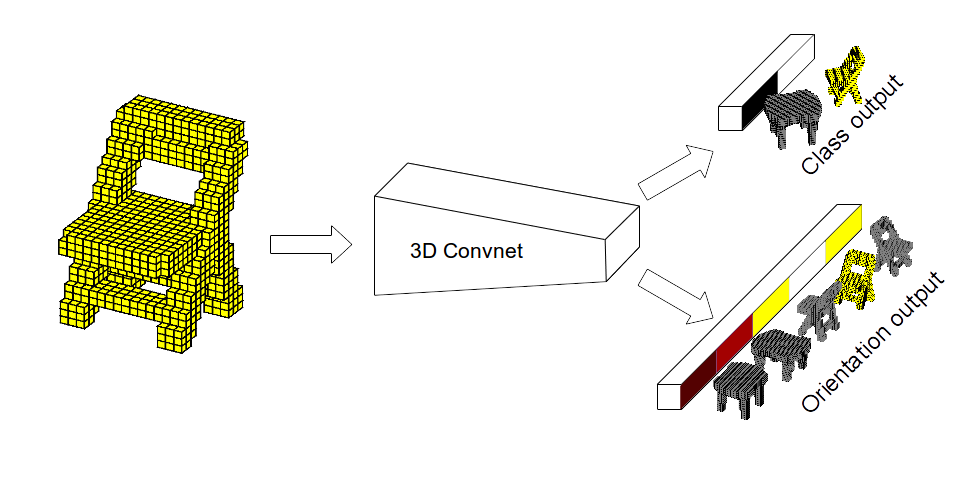}
  \end{center}
  \caption{Adding orientation classification as an auxiliary task to a 3D classification network improves its category-level classification accuracy.}
  \label{fig:teaser}
\end{figure}

In the present paper, we focus on the concept of object orientation. The actual task only cares about the object label, not its orientation. However, to produce the correct class label, at least some part of the network representation must be invariant to the orientation of the object, which is not trivial in 3D. Effectively, to be successful on the classification task, the network must also solve the orientation estimation task, but the loss function does not give any direct indication that solving this auxiliary task is important.
We show that forcing the network to produce the correct orientation during training increases its classification accuracy significantly -- Figure~\ref{fig:teaser}.

We introduce a network architecture that implements this idea and evaluate it on 4 different datasets representing the large variety of acquisition methods for point clouds: laser range scanners, RGB-D images, and CAD models.
The input to the network is an object candidate obtained from any of these data sources, which is fed to the network as an occupancy grid.
We compare the baseline without orientation information to our orientation-boosted version and obtain improved results in all experiments. We also compare to the existing 3D classification methods and achieve state-of-the-art results using our \textit{shalow} orientation-boosted networks in most of the experiments.
In the scope of our experiments, we extended the Modelnet40 dataset, which consists of more than 12k objects, with per-class alignments by using some automated alignment procedure~\cite{Sedaghat2015}. We will provide the additional annotation. 

We also applied the classifier in a 3D detection scenario using a simple 3D sliding box approach. In this context, the orientation estimation is no longer just an auxiliary task but also determines the orientation of the box, which largely reduces the runtime of the 3D detector. 

\section{Related Work}
Most previous works on 3D object recognition rely on handcrafted feature descriptors, such as
Point Feature Histograms \cite{rusu_learning_2008,Rusu2009}, 3D Shape Context \cite{Kortgen2003a}, or Spin Images \cite{johnson1999using}. Descriptors based on surface normals have been very popular, too \cite{Horn1984,PattersonIV2008}. Yulanguo~\etal~\cite{yulanguo_3d_2014} gives an extensive survey on such descriptors.

Feature learning for 3D recognition has first appeared in the context of RGB-D images, where depth is treated as an additional input channel \cite{farabet_learning_2013,couprie_indoor_2013,bo_unsupervised_2013}.
Thus, the approaches are conceptually very similar to feature learning in images.
Gupta~\etal~\cite{gupta_aligning_2015} fits and projects 3D synthetic models into the image plane.

3D convolutional neural networks (CNNs) have appeared in the context of videos. Tran \etal \cite{tran_c3d_2014} use video frame stacks as a 3D signal to approach multiple video classification tasks using their 3D CNN, called C3D. 3D CNNs are not limited to videos, but can be applied also to other three-dimensional inputs, such as point clouds, as in our work.

The most closely related works are by Wu~\etal~\cite{wu_3D_2015} and Maturana \& Sherer \cite{maturana_voxnet_2015}, namely 3D ShapeNets and VoxNet. Wu \etal use a Deep Belief Network to represent geometric 3D shapes as a probability distribution of binary variables on a 3D voxel grid. They use their method for shape completion from depth maps, too. The ModelNet dataset was introduced along with their work. The VoxNet \cite{maturana_voxnet_2015} is composed of a simple but effective CNN, accepting as input voxel grids similar to Wu \etal \cite{wu_3D_2015}. In both of these works the training data is augmented by rotating the object to make the network learn a feature representation that is invariant to rotation. However, in contrast to the networks proposed in this paper, the network is not enforced to output the object orientation, but only its class label. While in principle, the loss on the class label alone should be sufficient motivation for the network to learn an invariant representation, our experiments show that an explicit loss on the orientation helps the network to learn such representation.

Su et al.~\cite{su_multi-view_2015} take advantage of the object pose explicitly by rendering the 3D objects from multiple viewpoints and using the projected images in a combined architecture of 2D CNNs to extract features. However, this method still relies on the appearance of the objects in images, which only works well for dense surfaces that can be rendered. For sparse and potentially incomplete point clouds, the approach is not applicable. Song et al.~\cite{song_deep_2015} focus on 3D object detection in RGB-D scenes. They utilize a 3D CNN for 3D object bounding box suggestion. For the recognition part, they combine geometric features in 3D and color features in 2D.

Several 3D datasets have become available recently. Sydney Urban Objects~\cite{dedeuge_unsupervised_2013} includes point-clouds obtained from range-scanners. SUN-RGBD~\cite{song2015sun} and SUN-3D~\cite{xiao2013sun3d} gather some reconstructed point-clouds and RGB-D datasets in one place and in some cases they also add extra annotations to the original datasets. We use the annotations provided for NYU-Depth V2 dataset~\cite{silberman_indoor_2012-1} by SUN-RGBD in this work. ModelNet is a dataset consisting of synthetic 3D object models  \cite{wu_3D_2015}. Sedaghat \& Brox \cite{Sedaghat2015} created a dataset of annotated 3D point-clouds of cars from monocular videos using structure from motion and some assumptions about the scene structure.

Quite recently and parallel to this work, there have been several works utilizing 2D or 3D CNNs merely on 3D CAD models of ModelNet~\cite{wu_3D_2015} in supervised \cite{hegde_fusionnet_2016,brock_generative_2016,garcia_garcia_pointnet_2016}, semi-supervised~\cite{ravanbakhsh_deep_2016} and unsupervised~\cite{wu_learning_2016} fashions. Although our main architecture is rather shallow compared to most of them, and we use a rather low resolution compared to methods relying on high-resolution input images, we still provide state-of-the-art results on the aligned ModelNet10 subset, and on-par results on a roughly and automatically aligned version of the ModelNet40 subset.

Most of the published works on detection try to detect objects directly in the 2D space of the image. Wang \& Posner~\cite{wang2015voting} were among the first ones to utilize point clouds to obtain object proposals. In another line of work Gonzalez~\etal~\cite{gonzalez_multiview_2015} and Chen~\etal~\cite{chen_multi-view_2016} mix both 2D and 3D data for detection. Many authors, including Li~\etal~\cite{li_deep_2016} and Huang~\etal~\cite{huang_densebox_2015} approach the task with a multi-tasking method.

\section{Method}
\begin{figure*}[!t]
  \begin{center}
    \includegraphics[width=.95\textwidth]{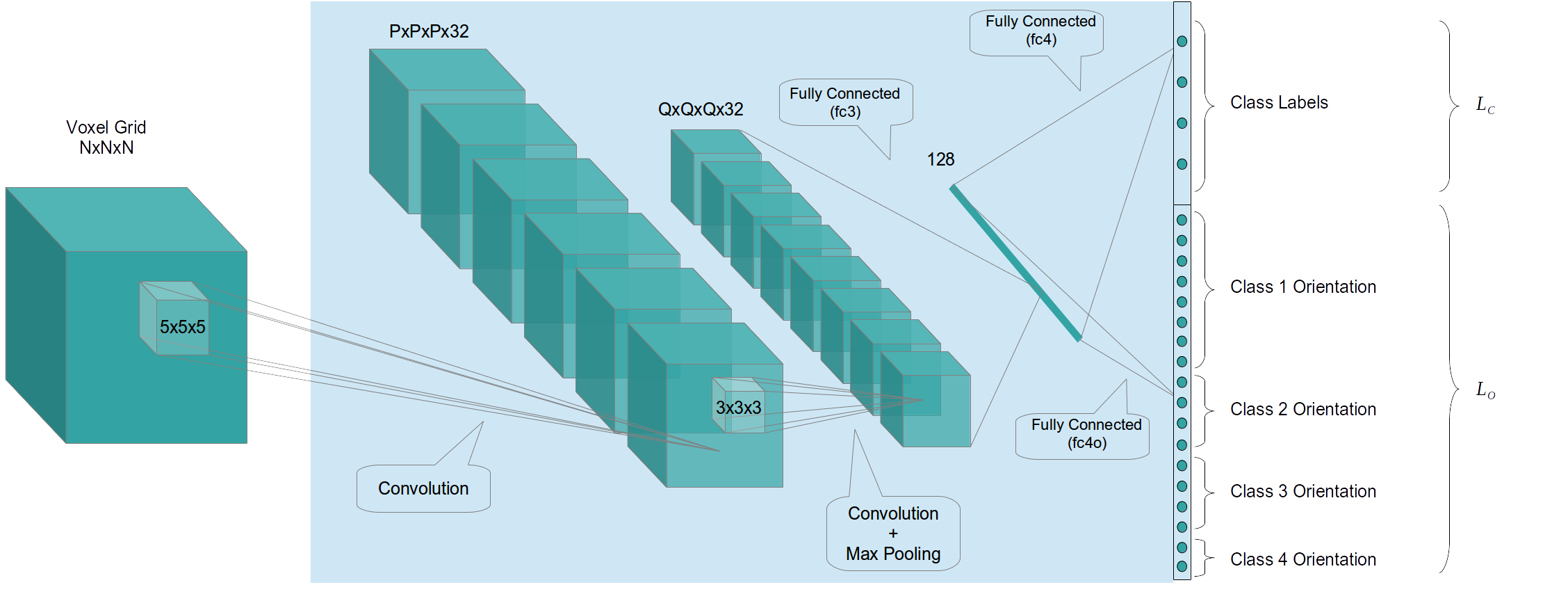}
  \end{center}
  \caption{Basic Orientation Boosting. Class labels and orientation labels are two separate outputs. The number of orientation labels assigned to each class can be different to the others. Both outputs contribute to the training equally -- with the same weight.}
  \label{fig:net1}
\end{figure*}

The core network architecture is based on VoxNet~\cite{maturana_voxnet_2015} and is illustrated in Fig.~\ref{fig:net1}. It takes a 3D voxel grid as input and contains two convolutional layers with 3D filters followed by two fully connected layers. Although this choice may not be optimal, we keep it to be able to directly compare our modifications to VoxNet. In addition, we experimented with a slightly deeper network that has four convolutional layers. 

Point clouds and CAD models are converted to voxel grids (occupancy grids). For the NYUv2 dataset we used the provided tools for the conversion; for the other datasets we implemented our own version. We tried both binary-valued and continuous-valued occupancy grids. In the end, the difference in the results was negligible and thus we only report the results of the former one.

\paragraph{Multi-task learning}We modify the baseline architecture by adding orientation estimation as an auxiliary parallel task. We call the resulting architecture the ORIentation-boosted vOxel Net -- ORION. Without loss of generality, we only consider rotation around the z-axis (azimuth) as the most varying component of orientation in practical applications. Throughout this paper we use the term 'orientation' to refer to this component.

Orientation is a continuous variable and the network could be trained to provide such an output. However, the idea is to treat different orientations of an object differently, and therefore we cast the orientation estimation as a classification problem. This also serves as a relaxation on dataset constraints, as a rough alignment of data obviates the need for strict orientation annotations. The network has output nodes for the product label space of classes and orientations and learns the mapping
\begin{equation}
\label{eq_map}
  x_i \mapsto (c_i,o_i)
\end{equation}

\noindent where $x_i$ are the input instances and $c_i$ , $o_i$ are their object class and \textit{orientation class}, respectively.

We do not put the same orientations from different object classes into the same orientation class, because we do not seek to extract any information from the absolute pose of the objects. Sharing the orientation output for all classes would make the network learn features shared among classes to determine the orientation, which is the opposite of what we want: leveraging the orientation estimation as an auxiliary task to improve on object classification. For example, a table from the $45\degree$ orientations class is not expected to share any useful information with a car of the same orientation. 

We choose multinomial cross-entropy losses \cite{rubinstein_cross-entropy_2013} for both tasks, so we can combine them by summing them up:
\begin{equation}
\label{eq_loss}
  \mathcal{L} = (1-\gamma)\mathcal{L}_C + \gamma\mathcal{L}_O
\end{equation}

\noindent where $\mathcal{L}_C$ and $\mathcal{L}_O$ indicate losses for object classification and orientation estimation tasks respectively. We used equal loss weights ($\gamma=0.5$) and found in our classification experiments that the results do not depend on the exact choice of the weight $\gamma$ around this value. However, in one of the detection experiments, where the orientation estimation is not an auxiliary task anymore, we used a higher weight for the orientation output to improve its accuracy.

Due to various object symmetries, the number of orientation labels differs per object class -- Figure~\ref{fig:net1}.
The idea is that we do not want the network to try to differentiate between, e.g., a table and its $180^{\circ}$ rotated counterpart. For the same reason, to rotationally symmetric objects, such as poles, or rotationally \emph{neutral} ones, such as trees to which no meaningful azimuth label can be assigned, we dedicate only a single node.
This is decided upon manually in the smaller datasets. However, during the auto-alignment of the bigger Modelnet40 dataset, the number of orientations are also automatically assigned to different classes. Details are given in the supplementary material.

\paragraph{Voting}
Object orientations can be leveraged at the entry of the network, too. During the test phase we feed multiple rotations of the test object to the network and obtain a final consensus on the class label based on the votes we obtain from each inference pass, as follows:

\begin{equation}
\label{eq_map}
  c_{final} = \argmax_k{\sum_r{S_k(x_r)}}
\end{equation}

\noindent where $S_k$ is the score the network assigns to the object at its $k^{th}$ node of the main (object category) output layer. $x_r$ is the test input with the rotation index $r$.

\section{Datasets}
\begin{figure*}[!t]
  \begin{center}
    \begin{minipage}{\linewidth}
      \begin{subfigure}[b]{0.5\linewidth} \centering
        \includegraphics[width=.7\linewidth,height=.5\linewidth]{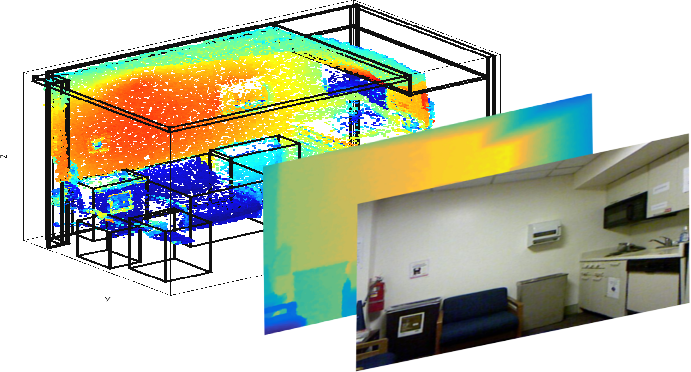}
      \end{subfigure}
      \begin{subfigure}[b]{.5\linewidth} \centering
        \includegraphics[width=.7\linewidth,height=.5\linewidth]{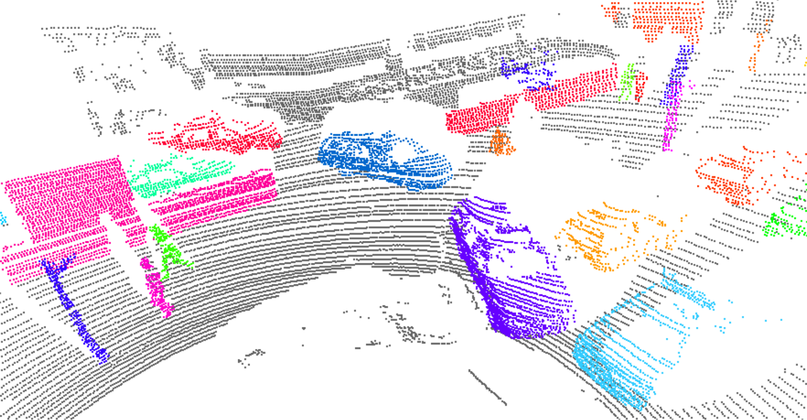}
      \end{subfigure}
    \end{minipage}
    \begin{subfigure}[b]{\linewidth} \centering
      \includegraphics[trim={0 2.5cm 0 0},clip,width=.85\linewidth]{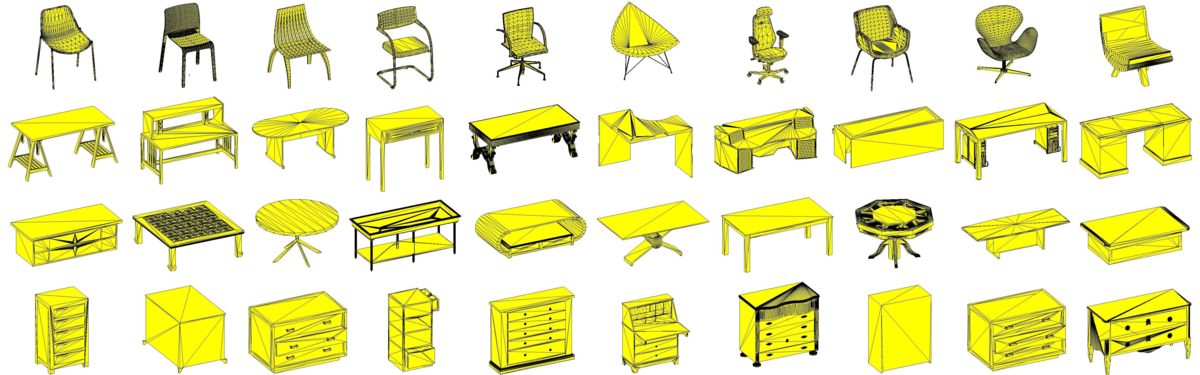}
    \end{subfigure}
  \end{center}
  \caption{Examples from the various 3D datasets we used in experiments. On top, two exemplar scenes from the NYUv2~\cite{silberman_indoor_2012-1} \& Sydney~\cite{dedeuge_unsupervised_2013} datasets are depicted. On the bottom samples from the Modelnet dataset are displayed. The KITTI dataset is similar to the Sydney dataset and is not shown here.}
  \label{fig:datasets}
\end{figure*}

We train and test our networks on four datasets, three of which are illustrated in Figure~\ref{fig:datasets}. We have chosen the datasets such that they represent different data sources.

\paragraph{Sydney Urban Objects - LiDAR/Pointcloud}
This dataset consists of LiDAR scans of 631 objects in 26 categories. The objects' point-clouds in this dataset are always incomplete, as they are only seen by the LiDAR sensor from a single viewpoint. Therefore the quality of the objects are by no means comparable to synthetic 3D objects, making classification a challenging task, even for human eyes; see Figure~\ref{fig:datasets}. This dataset is also of special interest in our category-level classification, as it provides a tough categorization of vehicles: \emph{4wd}, \emph{bus}, \emph{car}, \emph{truck}, \emph{ute} and \emph{van} are all distinct categories.
We use the same settings as \cite{maturana_voxnet_2015} to make our results comparable to theirs. The point clouds are converted to voxel-grids of size 32x32x32, in which the object occupies a 28x28x28 space. Zero-paddings of size 2 is used on each side to enable displacement augmentation during training. We also annotated the orientations to make the data suitable for our method. These we will provide to the public. 

\paragraph{NYUv2 - Kinect/RGBD}
This dataset consists of an overall number of 2808 RGBD images, corresponding to 10 object classes. The class types are shared with the ModelNet10 dataset. We used voxel grids of size 32x32x32, which contain the main object in the size of 28x28x28. The rest includes the context of the object and each object has a maximum number of 12 rotations. The dataset does not provide orientation annotations and therefore we used the annotations provided by the SUN-RGBD benchmark \cite{song2015sun}.

\paragraph{ModelNet - Synthetic/CAD}
This dataset is composed of synthetic CAD models. The ModelNet10 subset consists of uniformly aligned objects of the same classes as in the NYUv2 dataset. The object meshes in this dataset are converted to voxel grids of size 28x28x28, similar to the NYUv2 setting.
The ModelNet40 subset does not come with alignments (or orientation annotations). Thus, we provided manual annotation of orientation that we will make publicly available. In addition, we ran an unsupervised automated procedure to align the samples of ModelNet40. Please refer to supplemental material for details.

\paragraph{KITTI - LiDAR/Pointcloud}
The KITTI dataset \cite{Geiger2012CVPR} contains 7481 training images and 7518 test images in its object detection task. Each image represents a scene which also comes with a corresponding Velodyne point cloud. 2D and 3D bounding box annotations are provided in the images. Using the provided camera calibration parameters they can be converted into the coordinates of the Velodyne scanner. 
We use this dataset only in the detection experiment. To be able to report and analyze the effects of our method at multiple levels, we split the publicly available training set to 80\% and 20\% subsets for training and testing, respectively.

\section{Experiments and Results}
\subsection{Classification}
The classification results on all datasets are shown in Table~\ref{table:classification}.
For the Sydney Urban Objects dataset, we report the average F1 score weighted by class support as in \cite{wu_3D_2015} to be able to compare to their work. This weighted average takes into account that the classes in this dataset are unbalanced. For the other datasets we report the average accuracy. The Sydney dataset provides 4 folds/subsets to be used for cross-validation; in each experiment three folds are for training and one for testing. Also due to the small size of this dataset, we run each experiment three times with different random seeds, and report the average over all the 12 results.

\definecolor{LightGreen}{rgb}{0.9,0.9,0.9}
\newcommand{\clg}{\cellcolor{LightGreen}}
\renewcommand{\multirowsetup}{\centering} 
\setlength{\tabcolsep}{4pt}
\begin{table*}[t]
\small
  \begin{center}
    \begin{tabular}{clcccccc}
      \toprule

      \multicolumn{4}{c}{Method$\downarrow$} & {} &\multicolumn{3}{c}{Dataset} \\
      \cmidrule{1-4}
      \cmidrule{6-8}

      \multicolumn{2}{c}{} & {\# Conv} & {\# param} & {} & Sydney & NYUv2 &  ModelNet10\\

      \midrule
      \multirow{4}{*}{Hand-crafted feat.}

      &Recursive D~\cite{socher_recursive_2012}		    & - & - && -   		& 37.6		& -		\\
      &Recursive D+C~\cite{socher_recursive_2012}	    & - & - && -   		& 44.8		& -		\\
      &Triangle+SVM~\cite{dedeuge_unsupervised_2013}	& - & - && 67.1		& -		& -		\\
      &GFH+SVM~\cite{chen_performance_2014}		        & - & - && 71.0  	& -		& -		\\

      \cmidrule{2-8}
      \multirow{2}{*}{Deep Network}
      &FusionNet~\cite{hegde_fusionnet_2016}            &  & \normalsize{118M}  && -   	            & -	    & 93.1		\\
      &VRN\textsuperscript{\textdagger} \cite{brock_generative_2016}        		    & \normalsize{43} & \normalsize{18M} && -   	            & -	    & 93.6		\\
      \cmidrule{2-8}
      \multirow{5}{*}{Shallow Network}
      &ShapeNet~\cite{wu_3D_2015}        		        & 3 & - && -   		        & 57.9	& 83.5		\\
      &DeepPano~\cite{shi_deeppano_2015}        		& 4 & - && -   	            & -	    & 85.5		\\
      &VoxNet~\cite{maturana_voxnet_2015} (baseline)	& 2 &  890K && 72       		&  71  	& 92\\
      \cmidrule{3-8}
      &\multirow{2}{*}{ORION (Ours)}					    
                                                        & \clg 2  & \clg 910K & \clg & \clg \textbf{77.8}	& \clg \textbf{75.4}    & \clg \textbf{93.8}\\
      &                                                 & 4  & 4M  && 77.5	& \textbf{75.5}    & \textbf{93.9}\\

      \bottomrule
    \end{tabular}
  \end{center}
  \caption{Classification results and comparison to the state-of-the-art on three datasets. We report the overall classification accuracy, except for the Sydney dataset where we report the weighted average over F1 score. The auxiliary task on orientation estimation clearly improves the classification accuracy on all datasets.\textsuperscript{\textdagger} We report the single-network results for this method.}
  \label{table:classification}
\end{table*}

\newcommand*\clipgraph{\adjincludegraphics[width=.1\linewidth,trim={{.2\width} {.2\height} {.2\width} {.2\height}},clip]}
\renewcommand{\multirowsetup}{\centering} 
\newcolumntype{C}{>{\tiny\centering\arraybackslash}}
\setlength{\tabcolsep}{4pt}
\begin{figure}[h!]
\small
  \begin{center}
    \begin{tabular}{ccccccccc}
      {} &
      {\clipgraph{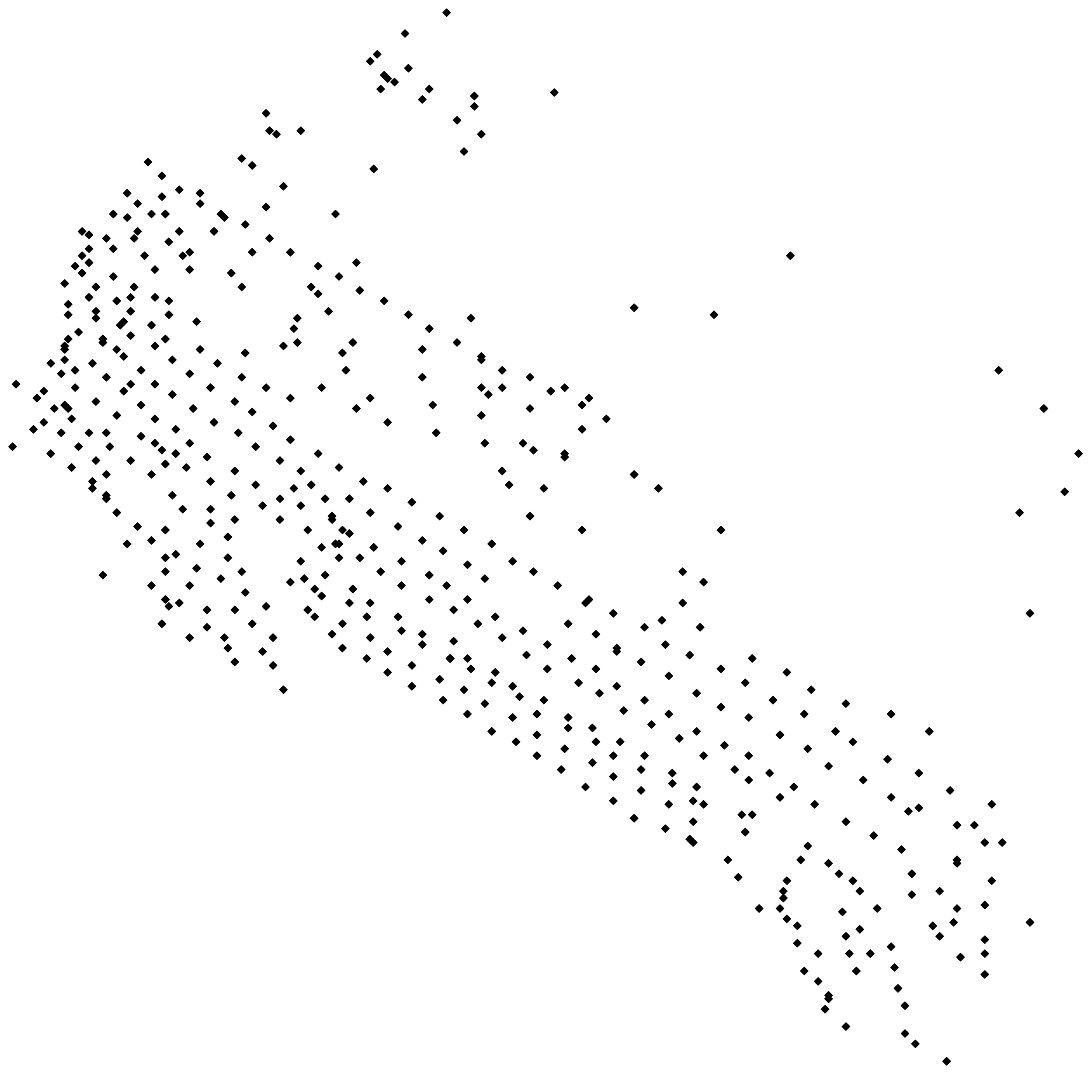}}&
      {\clipgraph{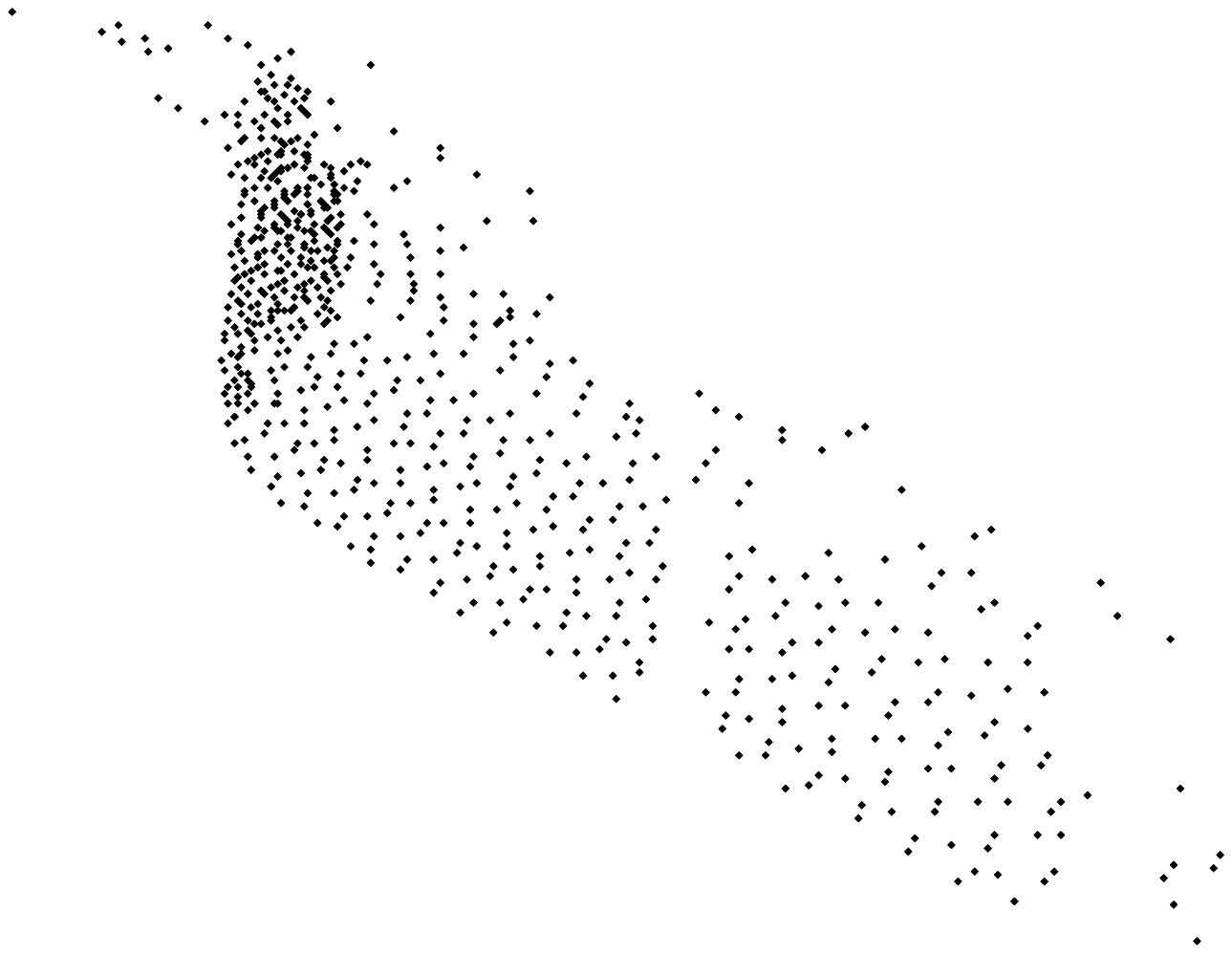}}&
      {\clipgraph{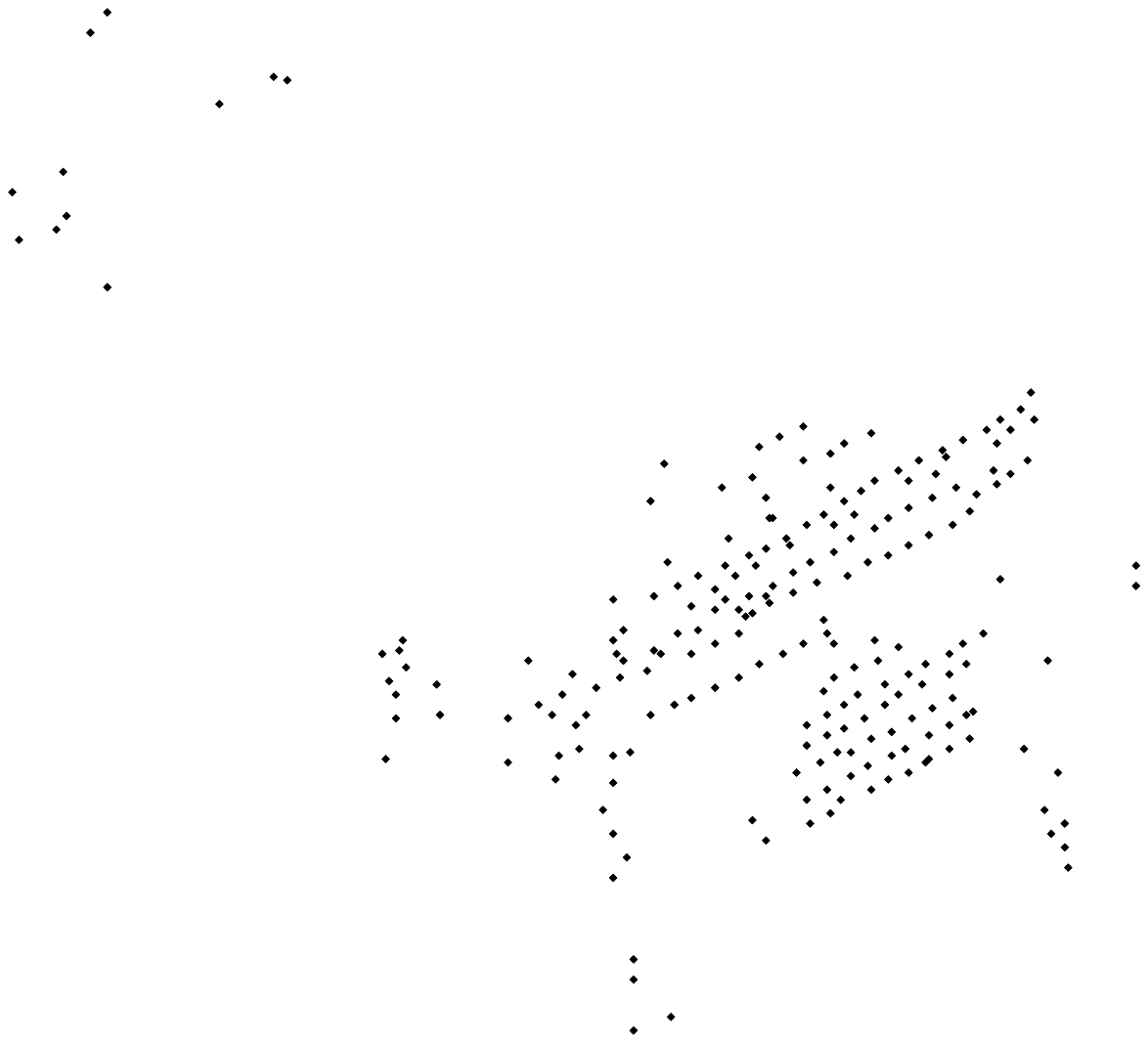}}&
      {\includegraphics[width=.1\linewidth]{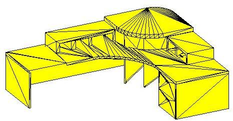}}&
      {\includegraphics[width=.1\linewidth]{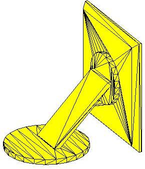}}&
      \\
      Ground Truth		  & 4wd 	& building 	& bus		& desk 		& monitor 	\\
      Baseline			  & \FA{car} 	& \FA{bus} 	& \FA{car} 	& \FA{sofa} 	& \FA{chair} 	\\
      Ours      		  & \TR{4wd}	& \TR{building}	& \TR{bus}	& \TR{desk} 	& \TR{monitor} 	\\
      {} &
      {\clipgraph{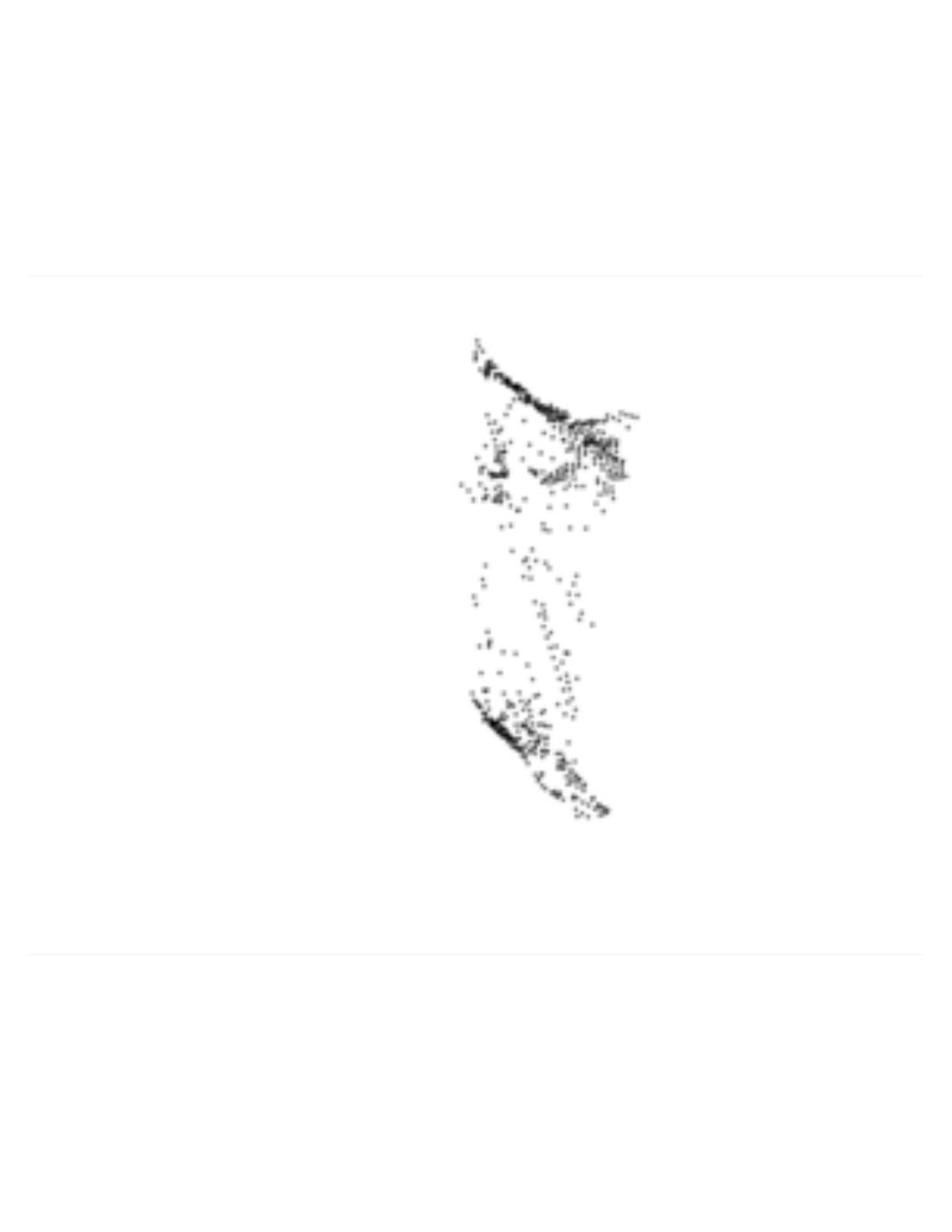}}&
      {\clipgraph{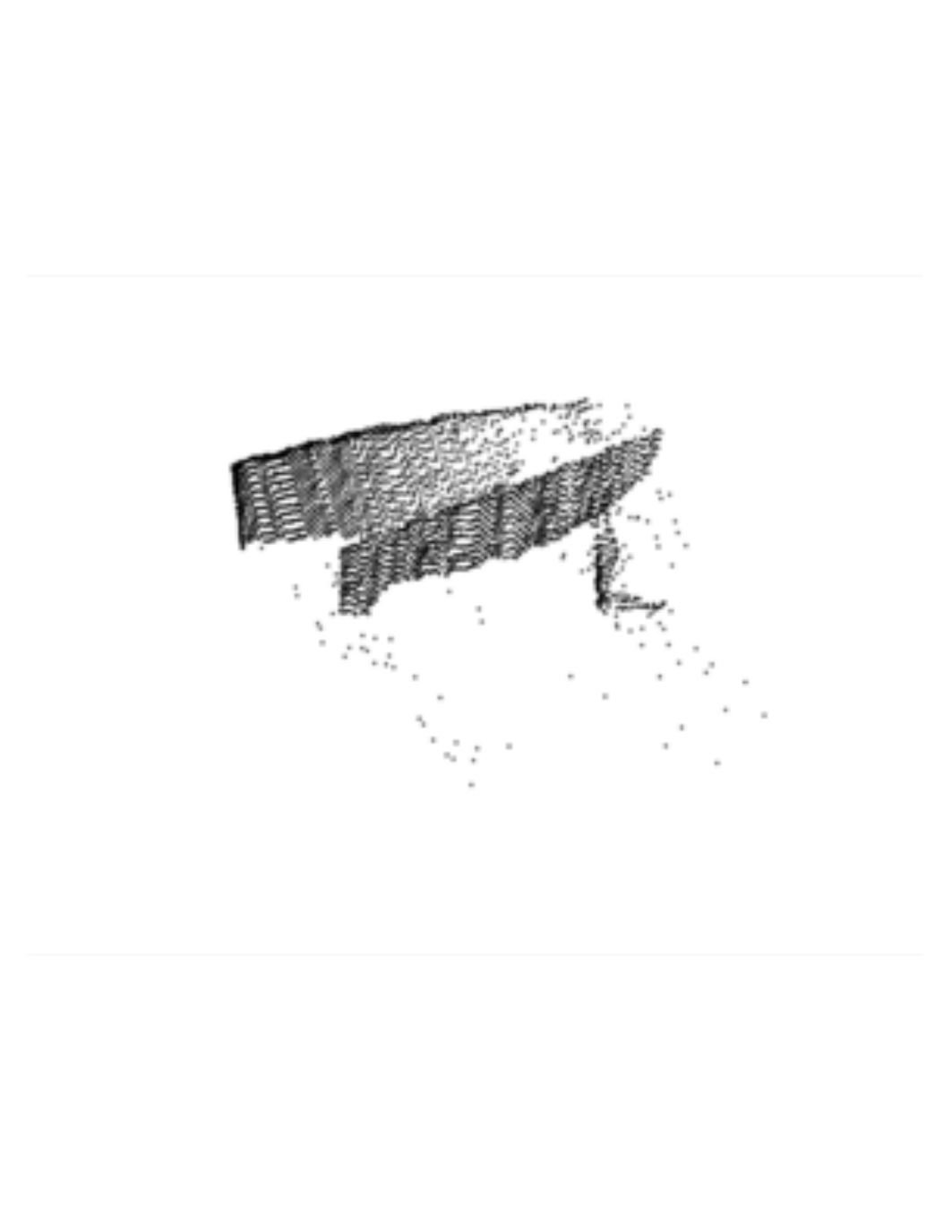}}&
      {\clipgraph{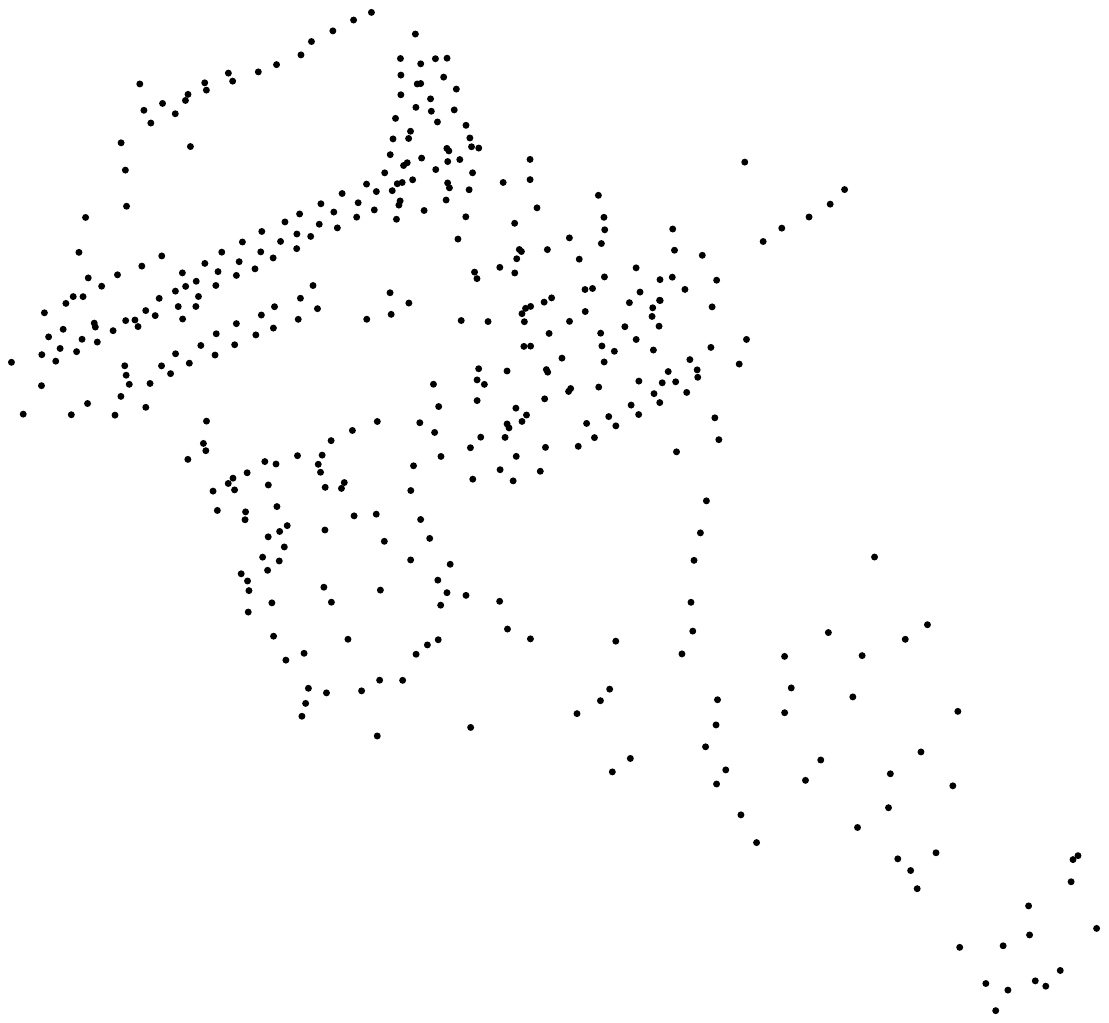}}&
      {\includegraphics[width=.1\linewidth]{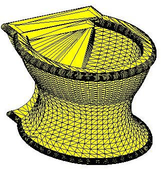}}&
      {\includegraphics[width=.1\linewidth]{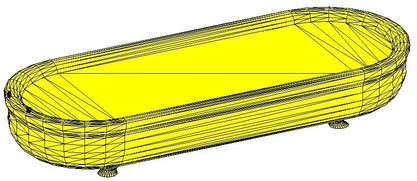}}&
      \\
      Ground Truth		  & table 	& chair	 	& ute		& toilet & bathtub\\
      Baseline			  & \FA{nite-stnd}& \FA{table} 	& \TR{ute}	& \TR{toilet} & \FA{table} \\
      Ours      		  & \TR{table}	& \TR{chair}	& \FA{truck}	& \FA{chair} & \FA{bed} \\
    \end{tabular}
  \end{center}
  \caption{Some exemplar classification results. We show examples on which the outputs of the two networks differ.}
  \label{fig:qualitative}
\end{figure}

We achieve clear improvements over the baseline and report state-of-the-art results in all the three datasets, with a far shallower architecture compared to previous state-of-the-art (2 vs. 43 conv. layers) and a big saving on number of parameters (1M vs.18M).

We also experimented with a slightly deeper network (last row of Table~\ref{table:classification}), but found that the network starts to overfit on the smaller datasets. Details of this extended architecture can be found in the supplementary material.

\renewcommand{\multirowsetup}{\centering} 
\setlength{\tabcolsep}{4pt}
\begin{table*}[t]
\small
  \begin{center}
    \begin{tabular}{lccccc}
      \toprule

      {} & {} & {} & \multicolumn{3}{c}{Accuracy (\%)} \\
      \cmidrule{4-6}

      {}        & {Conv.}   & {Batch}   & {No}          & {Rough, Automatic}    & {Perfect, Manual} \\
      {Method}  & {Layers}  & {Norm.}   & {Alignment}   & {Alignment}           & {Alignment} \\

      \midrule

      VoxNet~\cite{maturana_voxnet_2015} (baseline)	& 2 & $\times$      & 83 & - & -\\
      \cmidrule{2-6}
      \multirow{3}{*}{ORION (Ours)}					    
            					                    & 2 & $\times$      & -  & 88.1 & 87.5\\
         						                    & 2 & \checkmark 	& -  & 88.6 & 88.2\\
         						                    & 4 & \checkmark 	& -  & 89.4 & \textbf{89.7}\\

      \bottomrule
    \end{tabular}
  \end{center}
  \caption{Classification accuracy on Modelnet40. Orientation information during training clearly helps boost the classification accuracy even when orientation labels are obtained by unsupervised alignment \cite{Sedaghat2015}. In fact, manually assigned labels do not yield any significant improvement. Batch normalization and two additional convolutional layers improve results.}
  \label{table:modelnet40}
\end{table*}

\subsubsection{Non-aligned Dataset}
Since the Modelnet40 dataset does not come with alignments we manually annotated the orientation. As an alternative, we also aligned the objects, class by class, in an unsupervised fashion using the method introduced in Sedaghat \& Brox~\cite{Sedaghat2015}. Details of the steps of this process can be found in the supplementary material. Table \ref{table:modelnet40} shows the large improvement obtained by using the extra annotation during training. Interestingly, the automatic alignment is almost as good as the tedious manual orientation labeling. This shows that the network can benefit even from coarse annotations. 

Since the number of training samples is large, the deeper network with four convolutional layers performed even better than its counterpart with only two convolutional layers.  

Batch normalization (BN) is known to help with problems during network training \cite{icml2015_ioffe15}. 
Adding batch normalization to the convolutional layers yielded consistent improvements in the results; e.g. see Table~\ref{table:modelnet40}. We conjecture that batch normalization causes the error from the second task to propagate deeper back into the network.

\subsection{Detection}
We tested the performance of our suggested method in a detection scenario, where the orientation sensitive network is used as a binary object classifier to assign a score to 3D bounding box proposals in a sliding window fashion. We tested the 3D detector to detect cars in the KITTI dataset. 

\begin{figure}[!t]
    \begin{center}
        \begin{tabular}{cc}
            {\includegraphics[width=.45\textwidth]{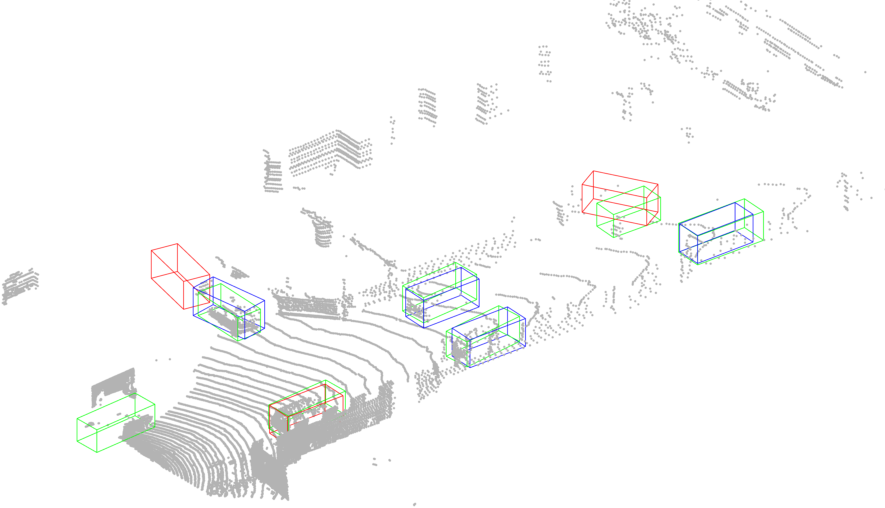}}
            &
            {\includegraphics[width=.45\textwidth]{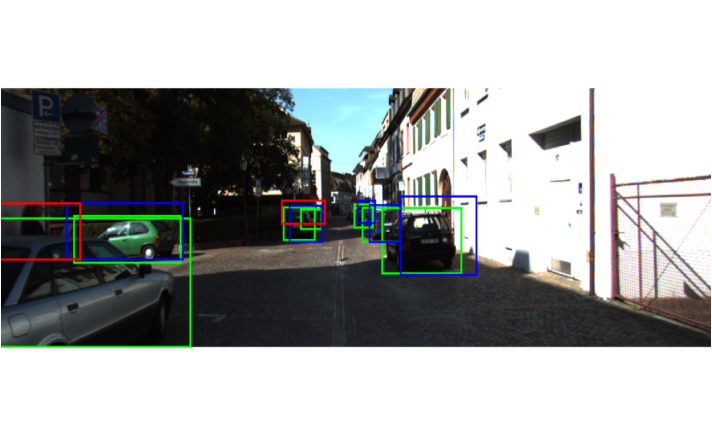}}
        \end{tabular}
        \vspace{-.5cm}
    \end{center}
  \begin{center}
    \includegraphics[width=.45\textwidth]{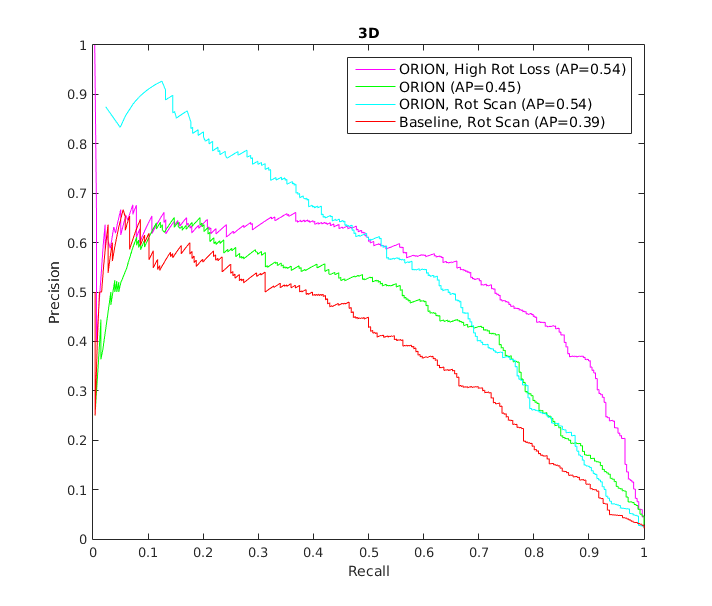}
    \includegraphics[width=.45\textwidth]{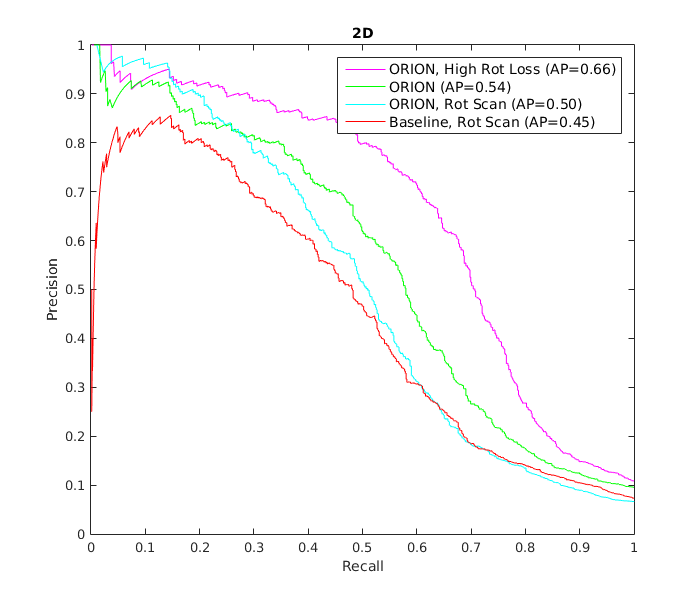}
  \end{center}
  \caption{On the top left, detected boxes of an exemplar scene are displayed in its 3D point-cloud representation. 3D boxes are then projected to the 2D image plane -- top right. Green boxes are the ground truth cars. Blue and red show true and false positives respectively. The bottom row illustrates the Precision-Recall curves for multiple detection experiments.}
  \label{fig:pr_curve}
\end{figure}

Figure~\ref{fig:pr_curve} quantifies the improvements of the ORION architecture in such an exemplar detection scenario. 
The mere use of our architecture as a binary classifier significantly pulls up the PR curve and increases the mean average precision. In this case, we only relied on the object classification output of the network and performed an exhaustive search over the rotations -- 18 rotation steps to cover 360 degrees. 
The main benefit is achieved when we also leveraged the orientation output of the network to directly predict the orientation of the object. This resulted in an $18\times$ run time improvement.
We also noticed that by increasing the loss weight of the orientation output, thus emphasizing on the orientation, the detection results improved further.

It is worth noting that in contrast to most of the detectors which run detection in the RGB image of the KITTI dataset, we do not use the RGB image but only use the 3D point cloud. We also limited the search in the scale and aspect-ratio spaces by obtaining statistical measures of the car sizes in the training set.

\section{Analysis}
To analyze the behavior of the orientation-boosted network, we compare it to its corresponding baseline network. To find a correspondence, we first train the baseline network long enough, so that it reaches a stable state. Then we use the trained net to initialize the weights of ORION, and continue training with low learning rate. We found that some filters tend to become more sensitive to orientation-specific features of the objects. We also found that in the baseline network some filters behave as the dominant ones for all the possible rotations of the objects in a class, while ORION has managed to spread the contributions over different filters for different orientations. Details of these experiments and visualizations are given in the supplementary material.

\section{Conclusions}
We showed for the task of 3D object classification that learning of certain concepts, such as invariance to object orientation, can be supported by adding the concept as an auxiliary task during training. By forcing the network to produce also the object orientation during training, it achieved better classification results at test time. This finding was consistent on all datasets and enabled us to establish state-of-the-art results on most of them. The approach is also applicable to 3D detection in a simple sliding 3D box fashion. In this case, the orientation output of the network avoids the exhaustive search over the object rotation. 

\section*{Acknowledgements}
We acknowledge funding by the ERC Starting Grant VideoLearn.
\bibliography{ref}

\begin{thebibliography}{40}
\providecommand{\natexlab}[1]{#1}
\providecommand{\url}[1]{\texttt{#1}}
\expandafter\ifx\csname urlstyle\endcsname\relax
  \providecommand{\doi}[1]{doi: #1}\else
  \providecommand{\doi}{doi: \begingroup \urlstyle{rm}\Url}\fi

\bibitem[Bo et~al.(2013)Bo, Ren, and Fox]{bo_unsupervised_2013}
Liefeng Bo, Xiaofeng Ren, and Dieter Fox.
\newblock Unsupervised feature learning for {{RGB-D}} based object recognition.
\newblock In \emph{Experimental {{Robotics}}}, pages 387--402. {Springer},
  2013.

\bibitem[Brock et~al.(2016)Brock, Lim, Ritchie, and
  Weston]{brock_generative_2016}
Andrew Brock, Theodore Lim, J.~M. Ritchie, and Nick Weston.
\newblock Generative and {{Discriminative Voxel Modeling}} with {{Convolutional
  Neural Networks}}.
\newblock \emph{arXiv:1608.04236 [cs, stat]}, August 2016.

\bibitem[Calakli and Taubin(2011)]{calakli2011ssd}
Fatih Calakli and Gabriel Taubin.
\newblock Ssd: Smooth signed distance surface reconstruction.
\newblock In \emph{Computer Graphics Forum}, volume~30, pages 1993--2002. Wiley
  Online Library, 2011.

\bibitem[Chen et~al.(2014)Chen, Dai, Liu, and Song]{chen_performance_2014}
Tongtong Chen, Bin Dai, Daxue Liu, and Jinze Song.
\newblock Performance of global descriptors for velodyne-based urban object
  recognition.
\newblock In \emph{Intelligent {{Vehicles Symposium Proceedings}}, 2014
  {{IEEE}}}, pages 667--673. {IEEE}, 2014.

\bibitem[Chen et~al.(2016)Chen, Ma, Wan, Li, and Xia]{chen_multi-view_2016}
Xiaozhi Chen, Huimin Ma, Ji~Wan, Bo~Li, and Tian Xia.
\newblock Multi-{{View 3D Object Detection Network}} for {{Autonomous
  Driving}}.
\newblock \emph{arXiv:1611.07759 [cs]}, November 2016.

\bibitem[Couprie et~al.(2013)Couprie, Farabet, Najman, and
  LeCun]{couprie_indoor_2013}
Camille Couprie, Cl{\'e}ment Farabet, Laurent Najman, and Yann LeCun.
\newblock Indoor semantic segmentation using depth information.
\newblock \emph{arXiv preprint arXiv:1301.3572}, 2013.

\bibitem[{De Deuge} et~al.(2013){De Deuge}, Robotics, Quadros, Hung, and
  Douillard]{dedeuge_unsupervised_2013}
Mark {De Deuge}, Field Robotics, Alastair Quadros, Calvin Hung, and Bertrand
  Douillard.
\newblock Unsupervised {{Feature Learning}} for {{Classification}} of {{Outdoor
  3D Scans}}.
\newblock 2013.

\bibitem[Farabet et~al.(2013)Farabet, Couprie, Najman, and
  LeCun]{farabet_learning_2013}
Clement Farabet, Camille Couprie, Laurent Najman, and Yann LeCun.
\newblock Learning hierarchical features for scene labeling.
\newblock \emph{Pattern Analysis and Machine Intelligence, IEEE Transactions
  on}, 35\penalty0 (8):\penalty0 1915--1929, 2013.

\bibitem[Garcia-Garcia et~al.(2016)Garcia-Garcia, Gomez-Donoso,
  Garcia-Rodriguez, Orts-Escolano, Cazorla, and
  Azorin-Lopez]{garcia_garcia_pointnet_2016}
A.~Garcia-Garcia, F.~Gomez-Donoso, J.~Garcia-Rodriguez, S.~Orts-Escolano,
  M.~Cazorla, and J.~Azorin-Lopez.
\newblock {{PointNet}}: {{A 3D Convolutional Neural Network}} for real-time
  object class recognition.
\newblock In \emph{2016 {{International Joint Conference}} on {{Neural
  Networks}} ({{IJCNN}})}, pages 1578--1584, July 2016.
\newblock \doi{10.1109/IJCNN.2016.7727386}.

\bibitem[Geiger et~al.(2012)Geiger, Lenz, and Urtasun]{Geiger2012CVPR}
Andreas Geiger, Philip Lenz, and Raquel Urtasun.
\newblock Are we ready for {{Autonomous Driving}}? {{The KITTI Vision Benchmark
  Suite}}.
\newblock In \emph{Conference on {{Computer Vision}} and {{Pattern
  Recognition}} ({{CVPR}})}, 2012.

\bibitem[Gonz{\'a}lez et~al.()Gonz{\'a}lez, Villalonga, Xu, V{\'a}zquez,
  Amores, and L{\'o}pez]{gonzalez_multiview_2015}
A.~Gonz{\'a}lez, G.~Villalonga, J.~Xu, D.~V{\'a}zquez, J.~Amores, and A.~M.
  L{\'o}pez.
\newblock Multiview random forest of local experts combining {{RGB}} and
  {{LIDAR}} data for pedestrian detection.
\newblock In \emph{2015 {{IEEE Intelligent Vehicles Symposium}} ({{IV}})},
  pages 356--361.
\newblock \doi{10.1109/IVS.2015.7225711}.

\bibitem[Gupta et~al.(2015)Gupta, Arbel{\'a}ez, Girshick, and
  Malik]{gupta_aligning_2015}
Saurabh Gupta, Pablo Arbel{\'a}ez, Ross Girshick, and Jitendra Malik.
\newblock Aligning {{3D Models}} to {{RGB-D Images}} of {{Cluttered Scenes}}.
\newblock In \emph{Proceedings of the {{IEEE Conference}} on {{Computer
  Vision}} and {{Pattern Recognition}}}, pages 4731--4740, 2015.

\bibitem[Hegde and Zadeh(2016)]{hegde_fusionnet_2016}
Vishakh Hegde and Reza Zadeh.
\newblock {{FusionNet}}: {{3D Object Classification Using Multiple Data
  Representations}}.
\newblock \emph{arXiv:1607.05695 [cs]}, July 2016.

\bibitem[Horn(1984)]{Horn1984}
Berthold Klaus~Paul Horn.
\newblock Extended gaussian images.
\newblock \emph{Proceedings of the IEEE}, 72\penalty0 (12):\penalty0
  1671--1686, 1984.

\bibitem[Huang et~al.(2015)Huang, Yang, Deng, and Yu]{huang_densebox_2015}
Lichao Huang, Yi~Yang, Yafeng Deng, and Yinan Yu.
\newblock {{DenseBox}}: {{Unifying Landmark Localization}} with {{End}} to
  {{End Object Detection}}.
\newblock \emph{arXiv:1509.04874 [cs]}, September 2015.

\bibitem[Ioffe and Szegedy(2015)]{icml2015_ioffe15}
Sergey Ioffe and Christian Szegedy.
\newblock Batch {{Normalization}}: {{Accelerating Deep Network Training}} by
  {{Reducing Internal Covariate Shift}}.
\newblock In David Blei and Francis Bach, editors, \emph{Proceedings of the
  32nd {{International Conference}} on {{Machine Learning}} ({{ICML}}-15)},
  pages 448--456. {JMLR Workshop and Conference Proceedings}, 2015.

\bibitem[Jacobson et~al.(2016)]{gptoolbox}
Alec Jacobson et~al.
\newblock {gptoolbox}: Geometry processing toolbox, 2016.
\newblock http://github.com/alecjacobson/gptoolbox.

\bibitem[Johnson and Hebert(1999)]{johnson1999using}
Andrew~E Johnson and Martial Hebert.
\newblock Using spin images for efficient object recognition in cluttered
  {{3D}} scenes.
\newblock \emph{Pattern Analysis and Machine Intelligence, IEEE Transactions
  on}, 21\penalty0 (5):\penalty0 433--449, 1999.

\bibitem[K{\"o}rtgen et~al.(2003)K{\"o}rtgen, Park, Novotni, and
  Klein]{Kortgen2003a}
Marcel K{\"o}rtgen, Gil-Joo Park, Marcin Novotni, and Reinhard Klein.
\newblock {{3D}} shape matching with {{3D}} shape contexts.
\newblock In \emph{The 7th central {{European}} seminar on computer graphics},
  volume~3, pages 5--17, 2003.

\bibitem[Li et~al.(2016)Li, Zia, Tran, Yu, Hager, and Chandraker]{li_deep_2016}
Chi Li, M.~Zeeshan Zia, Quoc-Huy Tran, Xiang Yu, Gregory~D. Hager, and Manmohan
  Chandraker.
\newblock Deep {{Supervision}} with {{Shape Concepts}} for
  {{Occlusion}}-{{Aware 3D Object Parsing}}.
\newblock \emph{arXiv:1612.02699 [cs]}, December 2016.

\bibitem[Maturana and Scherer(2015)]{maturana_voxnet_2015}
Daniel Maturana and Sebastian Scherer.
\newblock {{VoxNet}}: {{A 3D Convolutional Neural Network}} for {{Real-Time
  Object Recognition}}.
\newblock In \emph{Intelligent {{Robots}} and {{Systems}} ({{IROS}}), 2015
  {{IEEE}}/{{RSJ International Conference}} on}, pages 922--928. {IEEE}, 2015.

\bibitem[Miller(1994)]{miller_efficient_1994}
Gavin Miller.
\newblock Efficient {Algorithms} for {Local} and {Global} {Accessibility}
  {Shading}.
\newblock In \emph{Proceedings of the 21st {Annual} {Conference} on {Computer}
  {Graphics} and {Interactive} {Techniques}}, {SIGGRAPH} '94, pages 319--326,
  New York, NY, USA, 1994. ACM.
\newblock ISBN 978-0-89791-667-7.
\newblock \doi{10.1145/192161.192244}.
\newblock URL \url{http://doi.acm.org/10.1145/192161.192244}.

\bibitem[{Patterson IV} et~al.(2008){Patterson IV}, Mordohai, and
  Daniilidis]{PattersonIV2008}
Alexander {Patterson IV}, Philippos Mordohai, and Kostas Daniilidis.
\newblock Object detection from large-scale 3d datasets using bottom-up and
  top-down descriptors.
\newblock In \emph{Computer {{Vision{\textendash}ECCV}} 2008}, pages 553--566.
  {Springer}, 2008.
\newblock 00021.

\bibitem[Ravanbakhsh et~al.(2016)Ravanbakhsh, Schneider, and
  Poczos]{ravanbakhsh_deep_2016}
Siamak Ravanbakhsh, Jeff Schneider, and Barnabas Poczos.
\newblock Deep {{Learning}} with {{Sets}} and {{Point Clouds}}.
\newblock \emph{arXiv:1611.04500 [cs, stat]}, November 2016.

\bibitem[Rubinstein and Kroese()]{rubinstein_cross-entropy_2013}
Reuven~Y. Rubinstein and Dirk~P. Kroese.
\newblock \emph{The {{Cross}}-{{Entropy Method}}: {{A Unified Approach}} to
  {{Combinatorial Optimization}}, {{Monte}}-{{Carlo Simulation}} and {{Machine
  Learning}}}.
\newblock {Springer Science \& Business Media}.
\newblock ISBN 978-1-4757-4321-0.
\newblock Google-Books-ID: 8KgACAAAQBAJ.

\bibitem[Rusu et~al.(2008)Rusu, Marton, Blodow, and Beetz]{rusu_learning_2008}
Radu~Bogdan Rusu, Zoltan~Csaba Marton, Nico Blodow, and Michael Beetz.
\newblock Learning informative point classes for the acquisition of object
  model maps.
\newblock In \emph{Control, {{Automation}}, {{Robotics}} and {{Vision}}, 2008.
  {{ICARCV}} 2008. 10th {{International Conference}} on}, pages 643--650.
  {IEEE}, 2008.

\bibitem[Rusu et~al.(2009)Rusu, Blodow, and Beetz]{Rusu2009}
Radu~Bogdan Rusu, Nico Blodow, and Michael Beetz.
\newblock Fast point feature histograms ({{FPFH}}) for {{3D}} registration.
\newblock In \emph{Robotics and {{Automation}}, 2009. {{ICRA}}'09. {{IEEE
  International Conference}} on}, pages 3212--3217. {IEEE}, 2009.
\newblock 00311.

\bibitem[Sedaghat and Brox(2015)]{Sedaghat2015}
Nima Sedaghat and Thomas Brox.
\newblock Unsupervised {{Generation}} of a {{Viewpoint Annotated Car Dataset}}
  from {{Videos}}.
\newblock In \emph{Proceedings of the {{IEEE International Conference}} on
  {{Computer Vision}} ({{ICCV}})}, 2015.

\bibitem[Shi et~al.(2015)Shi, Bai, Zhou, and Bai]{shi_deeppano_2015}
Baoguang Shi, Song Bai, Zhichao Zhou, and Xiang Bai.
\newblock {{DeepPano}}: {{Deep Panoramic Representation}} for 3-{{D Shape
  Recognition}}.
\newblock \emph{IEEE Signal Processing Letters}, 22\penalty0 (12):\penalty0
  2339--2343, December 2015.
\newblock ISSN 1070-9908, 1558-2361.
\newblock \doi{10.1109/LSP.2015.2480802}.

\bibitem[Silberman et~al.(2012)Silberman, Hoiem, Kohli, and
  Fergus]{silberman_indoor_2012-1}
Nathan Silberman, Derek Hoiem, Pushmeet Kohli, and Rob Fergus.
\newblock Indoor segmentation and support inference from {{RGBD}} images.
\newblock In \emph{Computer {{Vision{\textendash}ECCV}} 2012}, pages 746--760.
  {Springer}, 2012.

\bibitem[Socher et~al.(2012)Socher, Huval, Bath, Manning, and
  Ng]{socher_recursive_2012}
Richard Socher, Brody Huval, Bharath Bath, Christopher~D. Manning, and
  Andrew~Y. Ng.
\newblock Convolutional-recursive deep learning for 3d object classification.
\newblock In \emph{Advances in {{Neural Information Processing Systems}}},
  pages 665--673, 2012.

\bibitem[Song and Xiao(2015)]{song_deep_2015}
Shuran Song and Jianxiong Xiao.
\newblock Deep {{Sliding Shapes}} for {{Amodal 3D Object Detection}} in {{RGB-D
  Images}}.
\newblock \emph{arXiv preprint arXiv:1511.02300}, 2015.

\bibitem[Song et~al.(2015)Song, Lichtenberg, and Xiao]{song2015sun}
Shuran Song, Samuel~P Lichtenberg, and Jianxiong Xiao.
\newblock {{SUN RGB-D}}: {{A RGB-D Scene Understanding Benchmark Suite}}.
\newblock In \emph{Proceedings of the {{IEEE Conference}} on {{Computer
  Vision}} and {{Pattern Recognition}}}, pages 567--576, 2015.

\bibitem[Su et~al.(2015)Su, Maji, Kalogerakis, and
  Learned-Miller]{su_multi-view_2015}
Hang Su, Subhransu Maji, Evangelos Kalogerakis, and Erik Learned-Miller.
\newblock Multi-view convolutional neural networks for {{3D}} shape
  recognition.
\newblock In \emph{Proceedings of the {{IEEE International Conference}} on
  {{Computer Vision}}}, pages 945--953, 2015.

\bibitem[Tran et~al.(2014)Tran, Bourdev, Fergus, Torresani, and
  Paluri]{tran_c3d_2014}
Du~Tran, Lubomir Bourdev, Rob Fergus, Lorenzo Torresani, and Manohar Paluri.
\newblock {{C3D}}: {{Generic Features}} for {{Video Analysis}}.
\newblock \emph{arXiv preprint arXiv:1412.0767}, 2014.

\bibitem[Wang and Posner()]{wang2015voting}
Dominic~Zeng Wang and Ingmar Posner.
\newblock Voting for {{Voting}} in {{Online Point Cloud Object Detection}}.
\newblock In \emph{Robotics: {{Science}} and {{Systems}}}.

\bibitem[Wu et~al.(2016)Wu, Zhang, Xue, Freeman, and
  Tenenbaum]{wu_learning_2016}
Jiajun Wu, Chengkai Zhang, Tianfan Xue, William~T Freeman, and Joshua~B
  Tenenbaum.
\newblock Learning a probabilistic latent space of object shapes via 3d
  generative-adversarial modeling.
\newblock In \emph{Advances in {{Neural Information Processing Systems}}},
  pages 82--90, 2016.

\bibitem[Wu et~al.(2015)Wu, Song, Khosla, Yu, Zhang, Tang, and
  Xiao]{wu_3D_2015}
Zhirong Wu, Shuran Song, Aditya Khosla, Fisher Yu, Linguang Zhang, Xiaoou Tang,
  and Jianxiong Xiao.
\newblock {{3D ShapeNets}}: {{A Deep Representation}} for {{Volumetric
  Shapes}}.
\newblock In \emph{Proceedings of the {{IEEE Conference}} on {{Computer
  Vision}} and {{Pattern Recognition}}}, pages 1912--1920, 2015.

\bibitem[Xiao et~al.(2013)Xiao, Owens, and Torralba]{xiao2013sun3d}
Jianxiong Xiao, Andrew Owens, and Antonio Torralba.
\newblock {{SUN3D}}: {{A}} database of big spaces reconstructed using sfm and
  object labels.
\newblock In \emph{Computer {{Vision}} ({{ICCV}}), 2013 {{IEEE International
  Conference}} on}, pages 1625--1632. {IEEE}, 2013.

\bibitem[{Yulan Guo} et~al.(2014){Yulan Guo}, Bennamoun, Sohel, {Min Lu}, and
  {Jianwei Wan}]{yulanguo_3d_2014}
{Yulan Guo}, Mohammed Bennamoun, Ferdous Sohel, {Min Lu}, and {Jianwei Wan}.
\newblock {{3D Object Recognition}} in {{Cluttered Scenes}} with {{Local
  Surface Features}}: {{A Survey}}.
\newblock \emph{IEEE Transactions on Pattern Analysis and Machine
  Intelligence}, 36\penalty0 (11):\penalty0 2270--2287, November 2014.
\newblock ISSN 0162-8828, 2160-9292.
\newblock \doi{10.1109/TPAMI.2014.2316828}.

\end{thebibliography}

\clearpage

\vspace*{1cm}
\noindent{\LARGE\bfseries\sffamily\textcolor{bmv@sectioncolor}{Supplementary Material}}
\vspace*{2cm}

\section{Auto-Alignment of the Modelnet40 dataset}

Modelnet40~\cite{wu_3D_2015} consists of more than 12000 \textit{non-aligned} objects in 40 classes. We used the method of Sedaghat \& Brox~\cite{Sedaghat2015} to automatically align the objects class by class.

\paragraph{Mesh to Point-Cloud Conversion}
The auto-alignment method of~\cite{Sedaghat2015} uses point-cloud representations of objects as input. Thus we converted the 3D mesh grids of Modelnet40 to point-clouds by assigning uniformly distributed points to object faces. 

Hidden faces in the mesh grids needed to be removed, as the so called Hierarchical Orientation Histogram (HOH) of \cite{Sedaghat2015} mainly relies on the exterior surfaces of the objects. We tackled this issue using the Jacobson's implementation~\cite{gptoolbox} of the ``ambient occlusion'' method \cite{miller_efficient_1994}.

We tried to distribute the points roughly with the same density across different faces, regardless of their shape and size, to avoid a bias towards bigger/wider ones.
Our basic point-clouds consist of around 50000 points per object, which are then converted to lighter models using the Smooth Signed Distance surface reconstruction method (SSD)~\cite{calakli2011ssd} as used in \cite{Sedaghat2015}.

\paragraph{Auto-Alignment}
We first created a ``reference-set'' in each class, consisting of a random subset of its objects, with an initial size of 100. This number was then decreased, as the low-quality objects were automatically removed from the reference set, according to \cite{Sedaghat2015}.
This reference set was then used to align the remaining objects of the class one by one.

For the HOH descriptor, we used 32 and 8 divisions in $\phi$ and $\theta$ dimensions respectively, for the root component. We also used 8 child components with 16 divisions for $\phi$ and 4 for $\theta$ -- see \cite{Sedaghat2015}.

\paragraph{Automatic Assignment of Number of Orientation Classes}
As pointed out in the main paper, we do not use the same number of orientation classes for all the object categories. We implemented the auto-alignment procedure in a way that this parameter is automatically decided upon for each category: During generation of the reference-set in each class, the alignment procedure was run with 3 different configurations, for which the search space spanned over 360, 180 and 90 degrees of rotations respectively. 
Each run resulted in an error measure representing the overall quality of the models selected as the reference-set, and we designated respectively 12, 6 and 3 orientation levels to each category, whenever possible. When none of these worked, e.g. for the 'flower_pot' class, we assigned 1 orientation class which is equivalent to discarding the orientation information.

\section{Analysis}
To analyze the behavior of the orientation-boosted network, we compare it to its corresponding baseline network. We would like to know the differences between corresponding filters in the two networks. To find this correspondence, we first train a baseline network, without orientations outputs, for long enough so that it reaches a stable state. Then we use this trained net to initialize the weights of the ORION network, and continue training with a low learning rate. This way we can monitor how the learned features change in the transition from the baseline to the orientation-aware network.

\begin{figure}[]
  \centering
  \begin{minipage}{.05\linewidth}
    \rotatebox{90}{ORION~$\xleftarrow{\makebox[10.3cm]{iterations}}$baseline}
  \end{minipage}
  \begin{minipage}{.9\linewidth}
    \begin{subfigure}[b]{.9\linewidth} \centering
      $\longleftarrow$~Object Orientations~$\longrightarrow$
    \end{subfigure}
    \begin{subfigure}[b]{.9\linewidth} \centering
      \fbox{\includegraphics[width=1\linewidth]{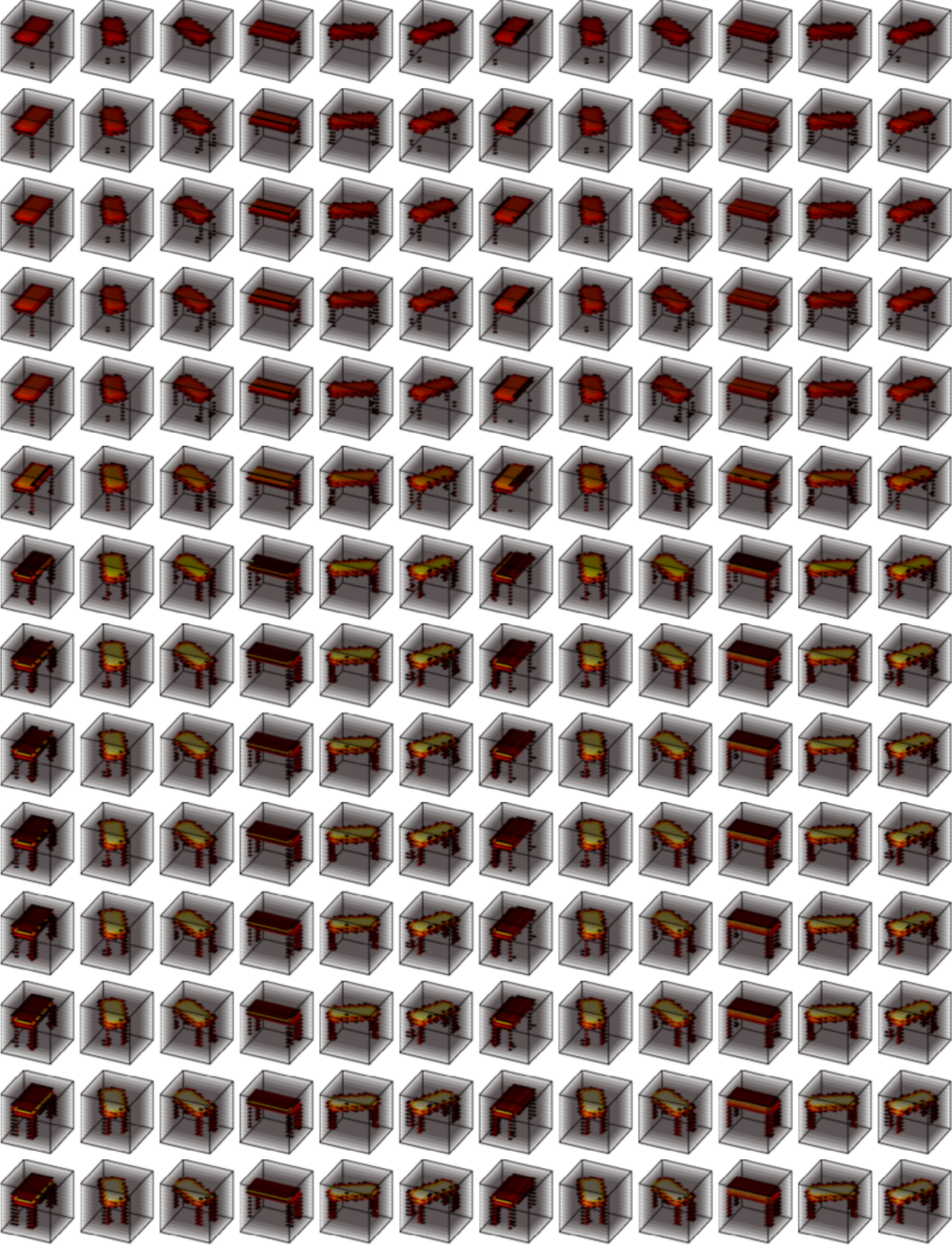}}
    \end{subfigure}
    \begin{subfigure}[b]{.9\linewidth} \centering
      \includegraphics[width=1\linewidth,height=0.2cm]{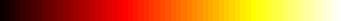}
    \end{subfigure}
    \begin{subfigure}[b]{.9\linewidth} \centering
      $\relbar$~higher values~$\longrightarrow$
    \end{subfigure}
  \end{minipage}
  \vspace*{.5cm}
  \caption{The picture illustrates the activations of one of the nodes of the first layer, while the network transitions from a baseline network to ORION. The input is always the same object, which has been fed to the network in its possible discretized rotations (columns) at each step (row). We simulated this transition by first training the baseline network and then fine-tuning our orientation-aware architecture on top of the learned weights. To be able to depict the 3D feature maps, we had to cut out values below a specific threshold. It can be seen that the encoded filter detects more orientation-specific aspects of the object, as it moves forward in learning the orientations. In addition, it seems that the filter is becoming more sensitive to a \emph{table} rather than only a horizontal surface -- notice the table legs appearing in the below rows.}
  \label{fig:featuremaps}
\end{figure}

In Figure~\ref{fig:featuremaps} transition of a single exemplar filter is depicted, and its responses to different rotations of an input object are illustrated. It turns out that the filter tends to become more sensitive to the orientation-specific features of the input object. Additionally some parts of the object, such as the table legs, show stronger response to the filter in the orientation-aware network.

\begin{figure}[]
  \centering

    \begin{subfigure}{\linewidth} \centering
      \makebox[1cm]{\rotatebox{90}{conv1}}\makebox[1cm]{\rotatebox{90}{conv2}}\makebox[1cm]{\rotatebox{90}{fc1}}\makebox[1cm]{\rotatebox{90}{fc2}}
      \\
      \includegraphics[width=.27\linewidth,height=.27\linewidth]{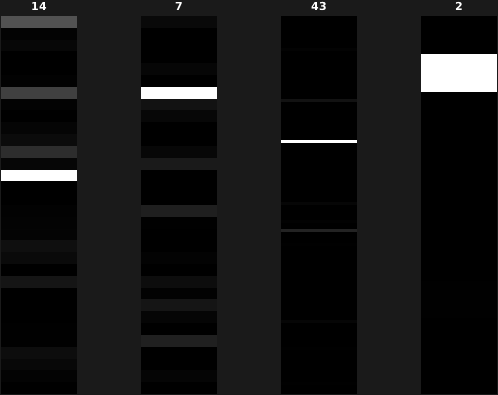}
      \caption{}
      \label{backtrace_big}
    \end{subfigure}

    \begin{subfigure}{\linewidth} \centering
      \rotatebox{90}{\tiny{baseline}}~\includegraphics[width=.9\linewidth,height=.07\linewidth]{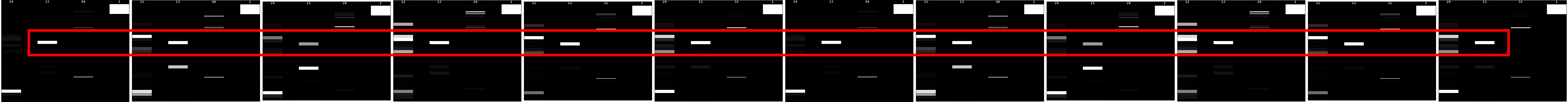}
      \rotatebox{90}{\tiny{ORION}}~\includegraphics[width=.9\linewidth,height=.07\linewidth]{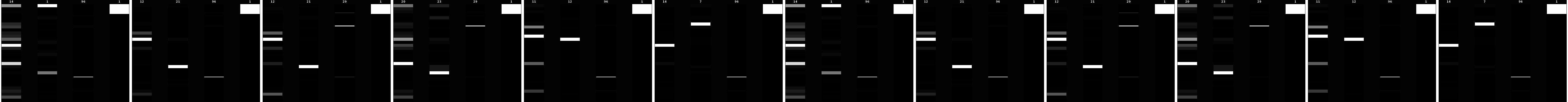}
      \\[1ex]
      \rotatebox{90}{\tiny{baseline}}~\includegraphics[width=.9\linewidth,height=.07\linewidth]{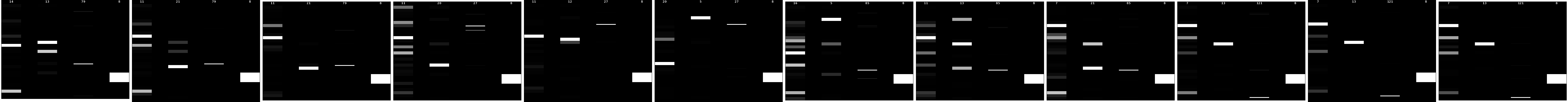}
      \rotatebox{90}{\tiny{ORION}}~\includegraphics[width=.9\linewidth,height=.07\linewidth]{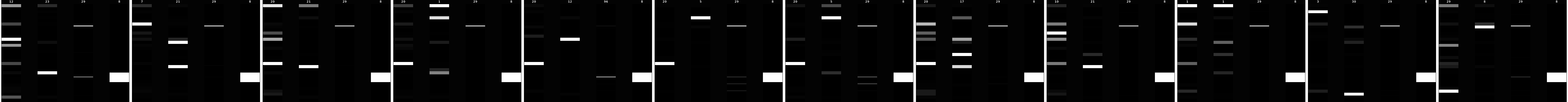}
      \\[1ex]
      \rotatebox{90}{\tiny{baseline}}~\includegraphics[width=.9\linewidth,height=.07\linewidth]{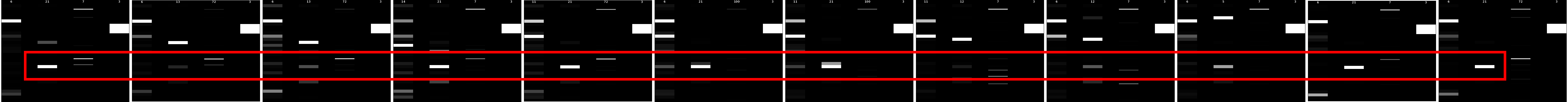}
      \rotatebox{90}{\tiny{ORION}}~\includegraphics[width=.9\linewidth,height=.07\linewidth]{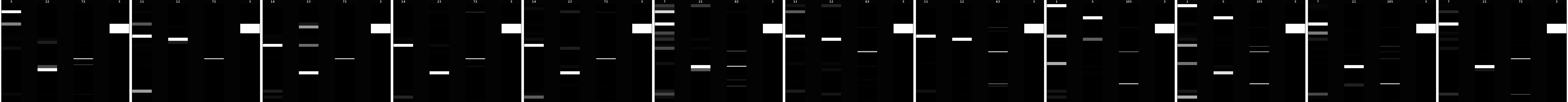}
      \\[1ex]      
      \rotatebox{90}{\tiny{baseline}}~\includegraphics[width=.9\linewidth,height=.07\linewidth]{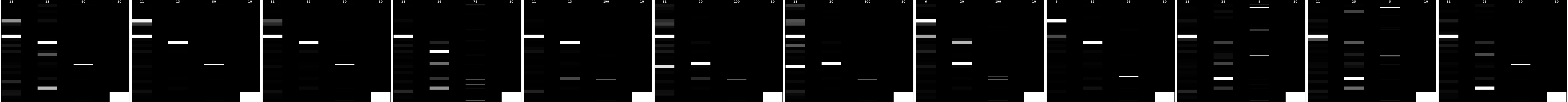}
      \rotatebox{90}{\tiny{ORION}}~\includegraphics[width=.9\linewidth,height=.07\linewidth]{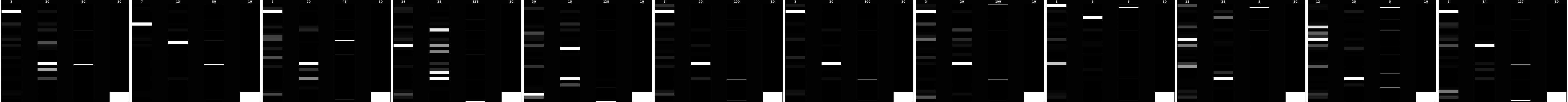}
      \\[1ex]
      \rotatebox{90}{\tiny{baseline}}~\includegraphics[width=.9\linewidth,height=.07\linewidth]{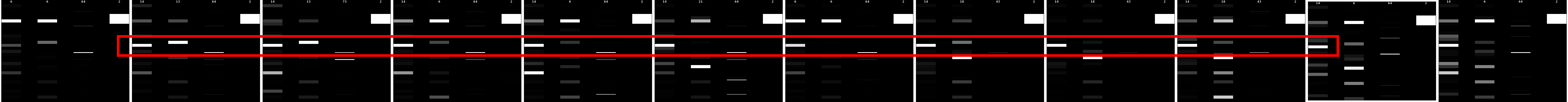}
      \rotatebox{90}{\tiny{ORION}}~\includegraphics[width=.9\linewidth,height=.07\linewidth]{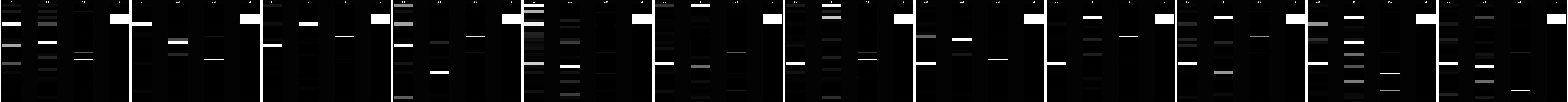}
      $\longleftarrow$~Object Orientations~$\longrightarrow$

      \caption{}
      \label{backtrace_matrix}
    \end{subfigure}
  \vspace*{.5cm}
  \caption{(a) shows the ``dominant signal-flow path'' of the network, for an exemplar object category-orientation. Each column contains the activations of one layer's nodes. Obviously the columns are of different sizes. Higher intensities show dominant nodes for the specific group of objects. Details of the steps taken to form such an illustration are explained in the text. In (b), rows represent object classes, while in different columns we show rotations of the objects. So each cell is a specific rotation of a specific object category. It can be seen that in the baseline network, many of the rotations of a class, share nodes in their dominant path (e.g. see the red boxes), whereas, in the ORION network the paths are more distributed over all the nodes.}
  \label{fig:backtrace}
\end{figure}

With such an observation, we tried to analyze the overall behavior of the network for specific object classes with different orientations. To this end we introduce the ``dominant signal-flow path'' of the network. The idea is that, although all the nodes and connections of the network contribute to the formation of the output, in some cases there may exist a set of nodes, which have a significantly higher effect in this process for an specific type of object/orientation. To test this, we take this step-by-step approach: First in a forward pass, the class, $c$, of the object is found. Then we seek to find the highest contributing node of the last hidden layer:

\begin{equation}
\label{eq_backtrace}
  l^{n-1}=\argmax_{k}\{{w^{n-1}_{k,c}a^{n-1}_k}\}
\end{equation}

where $n$ is the number of layers, $a^{n-1}_k$ are the activations of layer $n-1$, and $w^{n-1}_{k,c}$ is the weight connecting $a^{n-1}_k$ to the $c^{th}$ node of layer $n$. This way we naively assume there is a significant maximum in the \emph{contributions} and assign its index to $l^{n-1}$. Later we will see that this assumption proves to be true in many of our observations. We continue ``back-tracing'' the signal, to the previous layers. Extension of \eqref{eq_backtrace} to the convolutional layers is straight-forward, as we are just interested in finding the index of the node/filter in each layer. In the end, letting $l^n=c$, gives us the vector $l$ with length equal to the number of network layers, keeping the best contributors' indices in it.
Now to depict the ``dominant signal-flow path'' for a group of objects, we simply obtain $l$ for every member of the group, and plot the histogram of the $l^i$s as a column. Figure~\ref{fig:backtrace}(a) shows such an illustration for a specific class-rotation of the objects. It is clearly visible that for many objects of that group, specific nodes have been dominant.

In Figure~\ref{fig:backtrace}(b), the dominant paths of the baseline and ORION networks for some sample object categories of the Modelnet10 dataset are illustrated. It can be seen that in the baseline network, the dominant paths among various rotations of a class mostly share a specific set of nodes. This is mostly visible in the convolutional layers -- e.g. see the red boxes. On the contrary, the dominant paths in the ORION network rarely follow this rule and have more distributed path nodes. We interpret this as one of the results of orientation-boosting, and a helping factor in better classification abilities of the network.

\section{Extended Architecture}
\phantom{invisible text to make the table appear after section title}
\begin{table*}[h]
\small
  \begin{center}
    \begin{tabular}{lccccc}
      \toprule
      {} & {Conv1} & {Conv2} & {Conv3} & {Conv4} & {Pool4} \\
      \midrule
      \# of filters  & 32    & 64    & 128   & 256  & -     \\
      kernel size   & 3x3x3 & 3x3x3 & 3x3x3 & 3x3x3 & 2x2x2 \\
      stride        & 2     & 1     & 1     & 1     & 2     \\
      padding       & 0     & 0     & 0     & 0     & -     \\
      dropout ratio & 0.2   & 0.3   & 0.4   & 0.6   & -     \\
      batch normalization   & \checkmark   & \checkmark   & \checkmark  & \checkmark  & - \\
      \toprule
      {}            & {fc1} & {fc2:class} & {fc2:orientation}\\
      \midrule
      \# of outputs & 128   & 10/40    & variable\textsuperscript{\textdagger}  \\
      dropout ratio & 0.4   & -     & -  \\
      batch normalization & $\times$   & $\times$    & $\times$ \\
      \bottomrule
    \end{tabular}
  \end{center}
  \vspace*{.5cm}
  \caption{Details of the extended architecture introduced in Tables 1 \& 2 of the main article. \textsuperscript{\textdagger} The number of nodes dedicated to the orientation output varies in different experiments.}
  \label{table:ext_arch_details}
\end{table*}


\section{Orientation Estimation Results}
Although the orientation estimation was used merely as an auxiliary task, here in Table~\ref{table:orientation_results} we report the accuracies of the estimated orientation classes. Note that getting better results on orientation estimation would be possible by emphasizing on this task -- e.g. see the detection experiment in the main article.
\begin{table*}[h]
\small
  \begin{center}
    \begin{tabular}{lcccc}
      \toprule
      {}                & \multicolumn{4}{c}{Accuracy \%}\\
      \cmidrule{2-5}
      {}                & {Sydney}  & {NYUv2}   & {ModelNet10}  & {ModelNet40}\\
      \midrule
      ORION             & 71.5      & 51.9      & 89.0          & 86.5   \\
      ORION -- Extended & 70.1      & 54.5      & 89.3          & 87.6   \\
      \bottomrule
    \end{tabular}
  \end{center}
  \vspace*{.5cm}
  \caption{Orientation estimation accuracies on different datasets. The extended architecture of the second row, is the one introduced in the main article and detailed in Table~\ref{table:ext_arch_details} of this document.}
  \label{table:orientation_results}
\end{table*}


\end{document}